\documentclass{article}
\usepackage{ijcai16}

\usepackage{times}
\usepackage{amsmath}
\usepackage{amsthm}
\usepackage{amssymb}
\usepackage{enumitem}
\usepackage{graphicx}
\usepackage{fancyvrb}
\usepackage{color}
\definecolor{darkgreen}{rgb}{0,0.6,0}
\definecolor{darkblue}{rgb}{0,0,0.6}
\newtheorem{prop}{Proposition}
\newtheorem{thm}{Theorem}

\newtheorem{rmk}{Definition}

\usepackage{url}

\newcommand*{\QED}{\hfill\ensuremath{\square}}%

\title{Clustering Financial Time Series: How Long is Enough?}
\author{Gautier Marti \\ Hellebore Capital Ltd \\ Ecole Polytechnique
         \And
         S\'ebastien Andler \\ ENS de Lyon \\ Hellebore Capital Ltd
         \And
         Frank Nielsen \\ Ecole Polytechnique \\ LIX - UMR 7161
         \And
         Philippe Donnat \\ Hellebore Capital Ltd \\ Michelin House, London}

\begin{document}

\maketitle

\begin{abstract}
Researchers have used from 30 days to several years of daily returns as source data for clustering financial time series based on their correlations.
This paper sets up a statistical framework to study the validity of such practices.
We first show that clustering correlated random variables from their observed values is statistically consistent. Then, we also give a first empirical answer to the much debated question: How long should the time series be? If too short, the clusters found can be spurious; if too long, dynamics can be smoothed out.
\end{abstract}

\section{Introduction}

Clustering can be informally described as the task of
grouping objects in subsets (also named clusters) in
such a way that objects in the same cluster are more
similar to each other than those in different clusters.
Since the clustering task is notably hard to formalize
\cite{kleinberg2003impossibility}, designing a clustering algorithm
that solves it perfectly in any cases seems farfetched.
However, under strong mathematical assumptions made on the data,
desirable properties such as statistical consistency, i.e. more
data means more accuracy and in the limit a perfect solution, have
been shown: 
Starting from Hartigan's proof of Single Linkage \cite{hartigan1981consistency} and Pollard's proof
of $k$-means consistency \cite{pollard1981strong} to recent work such as the consistency of 
spectral clustering \cite{von2008consistency}, or modified $k$-means \cite{terada2013strong,terada2014strong}. These research papers assume that $N$ data points are
independently sampled from an underlying probability distribution in dimension $T$ fixed.
Clusters can be seen as regions of high density.
They show that in the large sample limit, $N \rightarrow \infty$, the clustering sequence constructed
by the given algorithm converges to a clustering of the whole underlying space.
When we consider the clustering of time series, another asymptotics matter: $N$ fixed and $T \rightarrow \infty$. Clusters gather objects that behave similarly through time.
To the best of our knowledge, much fewer researchers have dealt with this asymptotics:
\cite{borysov2014asymptotics} show the consistency of three hierarchical clustering algorithms when dimension $T$ is growing to correctly gather $N = n+m$ observations from a mixture of two $T$ dimensional Gaussian distributions $\mathcal{N}(\mu_1,\sigma_1^2 I_T)$ and $\mathcal{N}(\mu_2,\sigma_2^2 I_T)$.
\cite{ryabko2010clustering,khaleghi2012online} prove the consistency of $k$-means for clustering processes according to their \textit{distribution}. In this work, motivated by the clustering of financial time series, we will instead consider the consistency of clustering $N$ random variables from their $T$ observations according to their observed \textit{correlations}.

For financial applications, clustering is usually used as a building block before further processing such as portfolio selection \cite{tola2008cluster}.
Before becoming a mainstream methodology among practitioners, one has to provide theoretical guarantees that the approach is sound. In this work, we first show that the clustering methodology is theoretically valid, but when working with finite length time series extra care should be taken: Convergence rates depend on many factors (underlying correlation structure, separation between clusters, underlying distribution of returns) and implementation choice (correlation coefficient, clustering algorithm).
Since financial time series are thought to be approximately stationary for short periods only, a clustering methodology that requires a large sample to recover the underlying clusters is unlikely to be useful in practice and can be misleading. In section~\ref{rates}, we illustrate on simulated time series the empirical convergence rates achieved by several clustering approaches.

\section*{Notations}

\begin{itemize}[noitemsep,nolistsep]
\item $X_1,\ldots,X_N$ univariate random variables
\item $X_i^t$ is the $t^{\mathrm{th}}$ observation of variable $X_i$
\item $X_i^{(t)}$ is the $t^{\mathrm{th}}$ sorted observation of $X_i$
\item $F_X$ is the cumulative distribution function of $X$
\item $\rho_{ij} = \rho(X_i,X_j)$ correlation between $X_i, X_j$
\item $d_{ij} = d(X_i,X_j)$ distance between $X_i, X_j$
\item $D_{ij} = D(C_i,C_j)$ distance between clusters $C_i, C_j$
\item $P_k = \{C^{(k)}_1,\ldots,C^{(k)}_{l_k}\}$ is a partition of $X_1,\ldots,X_N$
\item $C^{(k)}(X_i)$ denotes the cluster of $X_i$ in partition $P_k$
\item $\Vert \Sigma \Vert_\infty = \max_{ij} \Sigma_{ij}$
\item $X = O_p(k)$ means $X/k$ is stochastically bounded, i.e. $\forall \varepsilon > 0, \exists M > 0, P(|X/k| > M) < \varepsilon$.
\end{itemize}

\section{The Hierarchical Correlation Block Model}

\subsection{Stylized facts about financial time series}

Since the seminal work in \cite{mantegna1999hierarchical}, it has been verified several times for different markets (e.g. stocks, forex, credit default swaps \cite{DBLP:conf/icmla/MartiVDN15}) that price time series of traded assets have a hierarchical correlation structure.
Another well-known stylized fact is the non-Gaussianity of daily asset returns \cite{cont2001empirical}. These empirical properties motivate both the use of alternative correlation coefficients described in section~\ref{correls} and the definition of the Hierarchical Correlation Block Model (HCBM) presented in section~\ref{hcbm}.


\subsection{Dependence and correlation coefficients}\label{correls}

The most common correlation coefficient is the Pearson correlation coefficient defined by
\begin{equation}
\rho(X,Y) = \frac{\mathbf{E}[XY] - \mathbf{E}[X] \mathbf{E}[Y]}{\sqrt{\mathbf{E}[X^2] - \mathbf{E}[X]^2}\sqrt{\mathbf{E}[Y^2] - \mathbf{E}[Y]^2}}
\end{equation}
which can be estimated by
\begin{equation}
\hat{\rho}(X,Y) = \frac{\sum_{t=1}^T (X^t - \overline{X})(Y^t - \overline{Y})}{\sqrt{\sum_{t=1}^T \left(X^t - \overline{X}\right)^2}\sqrt{\sum_{t=1}^T \left(Y^t - \overline{Y}\right)^2}}
\end{equation}
where $\overline{X} = \frac{1}{T}\sum_{t=1}^T X^t$ is the empirical mean of $X$.
This coefficient suffers from several drawbacks: it only measures linear relationship between two variables; it is not robust to noise and may be undefined if the distribution of one of these variables have infinite second moment. More robust correlation coefficients are copula-based dependence measures such as Spearman's rho
\begin{eqnarray}
\rho_S(X,Y) & = & 12 \int_{0}^{1} \int_{0}^{1} C(u,v) du dv - 3 \\
& = & 12~\mathbf{E}\left[F_X(X), F_Y(Y) \right] - 3 \\
& = & \rho\left(F_X(X),F_Y(Y)\right)
\end{eqnarray}
and its statistical estimate
\begin{equation}
\hat{\rho}_S(X,Y) = 1 - \frac{6}{T(T^2 - 1)}\sum_{t=1}^T \left(X^{(t)} - Y^{(t)} \right)^2.
\end{equation}
These correlation coefficients are robust to noise (since rank statistics normalize outliers) and invariant to monotonous transformations of the random variables (since copula-based measures benefit from the probability integral transform $F_X(X) \sim \mathcal{U}[0,1]$).

\subsection{The HCBM model}\label{hcbm}

\begin{figure}[!ht]
\begin{center}
\includegraphics[scale=0.455]{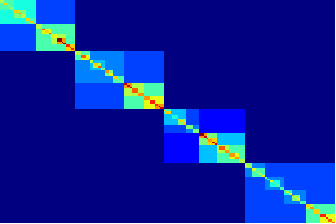}
\includegraphics[scale=0.455]{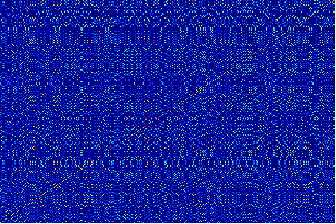}
\caption{(left) hierarchical correlation block model; (right) observed correlation matrix (following the HCBM) identical to the left one up to a permutation of the data}\label{fig:hbm}
\end{center}
\end{figure}

We assume that the $N$ univariate random variables $X_1,\ldots,X_N$ follow a Hierarchical Correlation Block Model (HCBM). This model consists in correlation matrices having a hierarchical block structure \cite{balakrishnan2011noise}, \cite{krishnamurthy2012efficient}. Each block corresponds to a correlation cluster that we want to recover with a clustering algorithm. In Fig.~\ref{fig:hbm}, we display a correlation matrix from the HCBM. Notice that in practice one does not observe the hierarchical block diagonal structure displayed in the left picture, but a correlation matrix similar to the one displayed in the right picture which is identical to the left one up to a permutation of the data.
The HCBM defines a set of nested partitions $\mathcal{P} = \{P_0 \supseteq P_1 \supseteq \ldots \supseteq P_h\}$ for some $h \in [1,N]$, where $P_0$ is the trivial partition, the partitions $P_k = \{C^{(k)}_1,\ldots,C^{(k)}_{l_k}\}$, and $\bigsqcup_{i = 1}^{l_k} C^{(k)}_i = \{X_1,\ldots,X_N\}$. For all $1 \leq k \leq h$, we define $\underline{\rho}_k$ and  $\overline{\rho}_k$ such that for all $1 \leq i,j \leq N$, we have $\underline{\rho}_k \leq \rho_{ij} \leq \overline{\rho}_k$ when $C^{(k)}(X_i) = C^{(k)}(X_j)$ and $C^{(k+1)}(X_i) \neq C^{(k+1)}(X_j)$, i.e. $\underline{\rho}_k$ and $\overline{\rho}_k$ are the minimum and maximum correlation respectively within all the clusters $ C^{(k)}_i$ in the partition $P_k$ at depth $k$. In order to have a proper nested correlation hierarchy, we must have $\overline{\rho}_k < \underline{\rho}_{k+1}$ for all $k$. Depending on the context, it can be a Spearman or Pearson correlation matrix.

Without loss of generality and for ease of demonstration we will consider the one-level HCBM with $K$ blocks of size $n_1,\ldots,n_K$ such that $\sum_{i=1}^K n_i = N$. We explain later how to extend the results to the general HCBM. 
We also consider the associated distance matrix $d$, where $d_{ij} = \frac{1-\rho_{ij}}{2}$.
In practice, clustering methods are applied on statistical estimates of the distance matrix $d$, i.e. on $\hat{d}_{ij} = d_{ij} + \epsilon_{ij}$, where $\epsilon_{ij}$ are noises resulting from the statistical estimation of correlations.

\section{Clustering methods}

\subsection{Algorithms of interest}

Many paradigms exist in the literature for clustering data. We consider in this work only hard (in opposition to soft) clustering methods, i.e. algorithms producing partitions of the data (in opposition to methods assigning several clusters to a given data point). 
Within the hard clustering family, we can classify for instance these algorithms in hierarchical clustering methods (yielding nested partitions of the data) and flat clustering methods (yielding a single partition) such as $k$-means.

We will consider the infinite Lance-Williams family which further subdivides the hierarchical clustering since many of the popular algorithms such as Single Linkage, Complete Linkage, Average Linkage (UPGMA), McQuitty's Linkage (WPGMA), Median Linkage (WPGMC), Centroid Linkage (UPGMC), and Ward's method are members of this family (cf. Table~\ref{LWparams} \cite{murtagh2012algorithms}). It will allow us a more concise and unified treatment of the consistency proofs for these algorithms. Interesting and recently designed hierarchical agglomerative clustering algorithms such as Hausdorff Linkage \cite{basalto2007hausdorff} and Minimax Linkage \cite{ao2005clustag} do not belong to this family \cite{bien2011hierarchical}, but their linkage functions share a convenient property for cluster separability.

\begin{table}
\centering
\caption{Many well-known hierarchical agglomerative clustering algorithms are members of the Lance-Williams family, i.e. the distance between clusters can be written: \newline $D(C_i \cup C_j, C_k) = \alpha_i D_{ik} + \alpha_j D_{jk} + \beta D_{ij} + \gamma |D_{ik} - D_{jk}|$}
\label{LWparams}
\begin{tabular}{l|c|c|c|}
\cline{2-4}
  & $\alpha_i$ & $\beta$ & $\gamma$   \\
  \hline
\multicolumn{1}{|l|}{Single} &  1/2  &  0      &     -1/2     \\
\hline
\multicolumn{1}{|l|}{Complete} &   1/2    &  0      &  1/2        \\
\hline
\multicolumn{1}{|l|}{Average}  &  $\frac{|C_i|}{|C_i| + |C_j|}$      &   0    &  0  \\
  \hline
\multicolumn{1}{|l|}{McQuitty}  &   1/2     &  0     & 0    \\
  \hline
\multicolumn{1}{|l|}{Median}  &   1/2     & -1/4      &  0  \\
  \hline
\multicolumn{1}{|l|}{Centroid}  &   $\frac{|C_i|}{|C_i| + |C_j|}$     & $- \frac{|C_i| |C_j|}{(|C_i|+ |C_j|)^2}$       &   0 \\
  \hline
\multicolumn{1}{|l|}{Ward}  &  $\frac{|C_i|+|C_k|}{|C_i|+|C_j|+|C_k|}$  & $- \frac{|C_k|}{|C_i|+|C_j|+|C_k|}$       &  0  \\
  \hline
\end{tabular}
\end{table}

\subsection{Separability conditions for clustering}

In our context the distances between the points we want to cluster are random and defined by the estimated correlations. However by definition of the HCBM, each point $X_i$ belongs to exactly one cluster $C^{(k)}(X_i)$ at a given depth $k$, and we want to know under which condition on the distance matrix we will find the correct clusters defined by $P_k$. We call these conditions the separability conditions.
A separability condition for the points $X_1,\ldots,X_N$ is a condition on the distance matrix of these points such that if we apply a clustering procedure whose input is the distance matrix, then the algorithm yields the correct clustering $P_k = \{C^{(k)}_1,\ldots,C^{(k)}_{l_k} \}$,  .
For example, for $\{X_1,X_2,X_3\}$ if we have $C(X_1) = C(X_2) \neq C(X_3)$ in the one-level two-block HCBM, then a separability condition is $d_{1,2} < d_{1,3}$ and $d_{1,2} < d_{2,3}$.

Separability conditions are deterministic and depend on the algorithm used for clustering. They are generic in the sense that for any sets of points that satisfy the condition the algorithm will separate them in the correct clusters. In the Lance-Williams algorithm framework \cite{chen1996space}, they are closely related to ``space conserving'' properties of the algorithm and in particular on the way the distances between clusters change during the clustering process.

\subsubsection{Space-conserving algorithms}

In \cite{chen1996space}, the authors define what they call a semi-space-conserving algorithm.

\begin{rmk}[Semi-space-conserving algorithms]
An algorithm is semi-space-conserving if for all clusters $C_i$, $C_j$, and $C_k$,
$$
 D(C_i\cup C_j,C_k) \in  \left[\min(D_{ik},D_{jk}), \max(D_{ik},D_{jk}) \right]
$$
\end{rmk}

Among the Lance-Williams algorithms we study here, Single, Complete, Average and McQuitty algorithms are semi-space-conserving. Although Chen and Van Ness only considered Lance-Williams algorithms the definition of a space conserving algorithm is useful for any agglomerative hierarchical algorithm. An alternative formulation of the semi-space-conserving property is:

\begin{rmk}[Space-conserving algorithms]
A linkage agglomerative hierarchical algorithm is space-conserving if $\displaystyle D_{ij}\in\left[\min_{x\in C_i,y\in C_j} d(x,y),\max_{x\in C_i,y\in C_j} d(x,y)\right]$.
\end{rmk}

Such an algorithm does not ``distort" the space when points are clustered which makes the sufficient separability condition easier to get. For these algorithms the separability condition does not depend on the size of the clusters.

The following two propositions are easy to verify.

\begin{prop}The semi-space-conserving Lance-Williams algorithms are space-conserving.
\end{prop}

\begin{prop} Minimax linkage and Hausdorff linkage are space-conserving.
\end{prop}

For space-conserving algorithms we can now state a sufficient separability condition on the  distance matrix.

\begin{prop}\label{spaceconservingcondition}
The following condition is a separability condition for space-conserving algorithms:
\begin{equation}
\tag{S1}
\max_{\substack{1 \leq i,j \leq N\\C(i) = C(j)}} d(X_i,X_j) ~< \min_{\substack{1 \leq i,j \leq N\\C(i) \neq C(j)}} d(X_i,X_j)
\label{sep_clust}
\end{equation}
The maximum distance is taken over any two points in a same cluster (intra) and the minimum over any two points in different clusters (inter).
\end{prop}

\proof Consider the set $\{d_{ij}^s\}$ of distances between clusters after $s$ steps of the clustering algorithm (therefore $\{d_{ij}^0\}$ is the initial set of distances between the points). Denote $\{d_{inter}^s\}$ (resp. $\{d_{intra}^s\}$) the sets of distances between subclusters belonging to different clusters (resp. the same cluster) at step $s$.
If the separability condition is satisfied then we have the following inequalities:

\begin{equation}
  \tag{S2}
\min d^0_{intra}\leq\max d^0_{intra}<\min d^0_{inter}\leq\max d^0_{inter}
\label{sep}
\end{equation}

Then the separability condition implies that the separability condition \ref{sep} is verified for all step $s$ because after each step the updated intra distances are in the convex hull of the intra distances of the previous step and the same is true for the inter distances. Moreover since \ref{sep} is verified after each step, the algorithm never links points from different clusters and the proposition entails. $\QED$

\subsubsection{Ward algorithm}
  
The Ward algorithm is a space-dilating Lance-Williams algorithm: $D(C_i\cup C_j,C_k)>\max(D_{ik},D_{jk})$. This is a more complicated situation because the structure $$\min d_{inter}<\max d_{inter}<\min d_{intra}<\max d_{intra}$$ is not necessarily preserved under the condition $\max d^0_{inter} < \min d^0_{intra}$. Points which are not clustered move away from the clustered points. Outliers, which will only be clustered at the very end, will end up close to each other and far from the clustered points. This can lead to wrong clusters. Therefore a generic separability condition for Ward needs to be stronger and account for the distortion of the space. Since the distortion depends on the number of steps the algorithm needs, the separability condition depends on the size of the clusters.

\begin{prop}[Separability condition for Ward]\label{ward_sepcond}
The separability condition for Ward reads:
$$n[\max d_{intra}^0-\min d_{intra}^0]< [\min d^0_{inter}-\min d^0_{intra}]$$
where $n = \max_i n_i$ is the size of the largest cluster.
\end{prop}
\proof Let $A$ and $B$ be two subsets of the $N$ points of size $a$ and $b$ respectively. Then
\begin{small}$$D(A,B)=\frac{ab}{a+b}\left(\frac{2}{ab}\sum_{\substack{i\in A\\j\in B}}d_{ij}-\frac{1}{a^2}\sum_{\substack{i\in A\\i'\in A}}d_{ii'}-\frac{1}{b^2}\sum_{\substack{j\in B\\j'\in B}}d_{jj'}\right)$$\end{small} is a linkage function for the Ward algorithm. 
To ensure that the Ward algorithm will never merge the wrong subsets it is sufficient that for any sets $A$ and $B$ in a same cluster, and $A'$, $B'$ in different clusters, we have:
$$D(A,B)<D(A',B').$$
Since
\begin{small}$$
\begin{cases}
D(A,B)\leq n(\max d^0_{intra}-\min d^0_{intra})+\min d^0_{intra}-1\\
D(A',B')\geq (\min d^0_{inter}-\max d^0_{intra})+\max d^0_{intra}-1
\end{cases}
$$\end{small}
we obtain the condition: 
$$n(\max d^0_{intra}-\min d^0_{intra})<\min d^0_{inter}-\min d^0_{intra}.$$
\QED

\subsubsection{$k$-means}

The $k$-means algorithm is not a linkage algorithm. 
For the $k$-means algorithm we need a separability condition that ensures that the initialization will be good enough for the algorithm to find the partition. In \cite{ryabko2010clustering} (Theorem 1), the author proves the consistency of the one-step farthest-point initialization $k$-means \cite{katsavounidis1994new} with a distributional distance for clustering processes. The separability condition \ref{sep_clust} of Proposition~\ref{spaceconservingcondition} is sufficient for $k$-means.

\section{Consistency of well-known clustering algorithms}

In the previous section we have determined configurations of points such that the clustering algorithm will find the right partition. The proof of the consistency now relies on showing that these configurations are likely. In fact the probability that our points fall in these configurations goes to $1$ as $T\to\infty$. 

The precise definition of what we mean by consistency of an algorithm is the following:

\begin{rmk}[Consistency of a clustering algorithm]
Let $(X_1^t,\ldots,X_N^t)$, $t=1,\ldots,T$, be $N$ univariate random variables observed $T$ times. A clustering algorithm $\mathcal{A}$ is consistent with respect to the Hierarchical Correlation Block Model (HCBM) defining a set of nested partitions $\mathcal{P}$ if the probability that the algorithm $\mathcal{A}$ recovers all the partitions in $\mathcal{P}$ converges to $1$ when $T \rightarrow \infty$.
\end{rmk}

As we have seen in the previous section the correct clustering can be ensured if the estimated correlation matrix verifies some separability condition. This condition can be guaranteed by requiring the error on each entry of the matrix $\hat{R}_T$ to be smaller than the contrast, i.e. $\frac{\underline{\rho}_1-\overline{\rho}_0}{2}$, on the theoretical matrix $R$. There are classical results on the concentration properties of estimated correlation matrices such as:

\begin{thm}[Concentration properties of the estimated correlation matrices \cite{liu2012high}]\label{liuthm}

If $\Sigma$ and $\hat{\Sigma}$ are the population and empirical Spearman correlation matrix respectively, then with probability at least $1-\frac{1}{T^2}$, for $N\geq \frac{24}{\log T}+2$, we have
$$\|\hat{\Sigma}-\Sigma\|_\infty\leq 24\sqrt{\frac{\log N}{T}}$$
\end{thm}

The concentration bounds entails that if $T \gg \log(N)$ then the clustering will find the correct partition because the clusters will be sufficiently separated with high probability. In financial applications of clustering, we need the error on the estimated correlation matrix to be small enough for relatively short time-windows. However there is a dimensional dependency of these bounds \cite{tropp2015introduction} that make them uninformative for realistic values of $N$ and $T$ in financial applications, but there is hope to improve the bounds using the special structure of HCBM correlation matrices.

\subsection{From the one-level to the general HCBM}

To go from the one-level HCBM to the general case we need to get a separability condition on the nested partition model. For both space-conserving algorithms and the Ward algorithm, this is done by requiring the corresponding separability condition for each level of the hierarchy. 

For all $1 \leq k \leq h$, we define $\underline{d}_k$ and $\overline{d}_k$ such that for all $1 \leq i,j \leq N$, we have $\underline{d}_k \leq d_{ij} \leq \overline{d}_k$ when $C^{(k)}(X_i) = C^{(k)}(X_j)$ and $C^{(k+1)}(X_i) \neq C^{(k+1)}(X_j)$. Notice that $\underline{d}_k = (1-\overline{\rho}_{k})/2$ and $\overline{d}_k = (1-\underline{\rho}_k)/2$.

\begin{prop}\label{nestedcondition}[Separability condition for space-conserving algorithms in the case of nested partitions] The separability condition reads:
$$\overline{d}_h < \underline{d}_{h-1} < \ldots < \overline{d}_{k+1} < \underline{d}_{k} < \ldots < \underline{d}_{1}.$$
\end{prop}

This condition can be guaranteed by requiring the error on each entry of the matrix $\hat{\Sigma}$ to be smaller than the lowest contrast.
Therefore the maximum error we can have for space-conserving algorithms on the correlation matrix is $$\Vert \Sigma-\hat{\Sigma}\Vert_\infty<\min_{k}\frac{\underline{\rho}_{k+1}-\overline{\rho}_k}{2}.$$

\begin{prop}\label{nested_ward}[Separability condition for the Ward algorithm in the case of nested partitions] Let $n_k$ be the size of the largest cluster at the level $k$ of the hierarchy.

The separability condition reads:
\begin{align*}
\forall k\in \{1,\ldots,h\}, & & n_k(\overline{d}_k-\underline{d}_h)< \underline{d}_{k-1}-\underline{d}_h
\end{align*}
\end{prop}

Therefore the maximum error we can have for space-conserving algorithms on the correlation matrix is $$\Vert \Sigma-\hat{\Sigma}\Vert_\infty<\min_{k}\frac{\overline{\rho}_{h}-\overline{\rho}_{k-1}-n_k(\overline{\rho}_h-\underline{\rho}_k)}{1+2n_k},$$
where $n_k$ is the size of the largest cluster at the level $k$ of the hierarchy.

We finally obtain consistency for the presented algorithms with respect to the HCBM from the previous concentration results.

\section{Empirical rates of convergence}\label{rates}

We have shown in the previous sections that clustering correlated random variables is consistent under the hierarchical correlation block model. This model is supported by many empirical studies \cite{mantegna1999hierarchical} where the authors scrutinize time series of returns for several asset classes. However, it was also noticed that the correlation structure is not fixed and tends to evolve through time.
This is why, besides being consistent, the convergence of the methodology needs to be fast enough for the underlying clustering to be accurate. For now, theoretical bounds such as the ones obtained in Theorem~\ref{liuthm} are uninformative for realistic values of $N$ and $T$. For example, for $N = 265$ and $T = 2500$ (roughly 10 years of historical daily returns) with a separation between clusters of $d = 0.2$, we are confident with probability greater than $1 - 2N^2 e^{-Td^2/24} \approx -2176$ that the clustering algorithm has recovered the correct clusters. These bounds will eventually converge to $0$ with rate $O_P(\sqrt{\log N} / \sqrt{T})$.
In addition, the convergence rates also depend on many factors, e.g. the number of clusters, their relative sizes, their separations, whose influence is very specific to a given clustering algorithm, and thus difficult to consider in a theoretical analysis. 

To get an idea of the minimal amount of data one should use in applications to be confident with the clustering results, we suggest to design realistic simulations of financial time series and determine the sample critical size from which the clustering approach ``always" recovers the underlying model. We illustrate
such an empirical study in the following section.

\subsection{Financial time series models}

For the simulations, implementation and tutorial available at \url{www.datagrapple.com/Tech}, we will consider two models:
\begin{itemize}
\item The standard but debated model of quantitative finance, the Gaussian random walk model whose increments are realizations from a $N$-variate Gaussian: $X \sim \mathcal{N}(0,\Sigma)$.
\end{itemize}
The Gaussian model does not generate heavy-tailed behavior (strong unexpected variations in the price of an asset) which can be found in many asset returns \cite{cont2001empirical} nor does it generate tail-dependence (strong variations tend to occur at the same time for several assets).
This oversimplified model provides an empirical convergence rate for clustering that is unlikely to be exceeded on real data.
\begin{itemize}
\item The increments are realizations from a $N$-variate Student's $t$-distribution, with degree of freedom $\nu = 3$: $X \sim t_{\nu}(0,\frac{\nu-2}{\nu}\Sigma)$.
\end{itemize}
The $N$-variate Student's $t$-distribution ($\nu = 3$) captures both the heavy-tailed behavior (since marginals are univariate Student's $t$-distribution with the same parameter $\nu = 3$) and the tail-dependence. It has been shown that this distribution yields a much better fit to real returns than the Gaussian distribution \cite{hu2010portfolio}.

\begin{figure}
\includegraphics[width=\columnwidth]{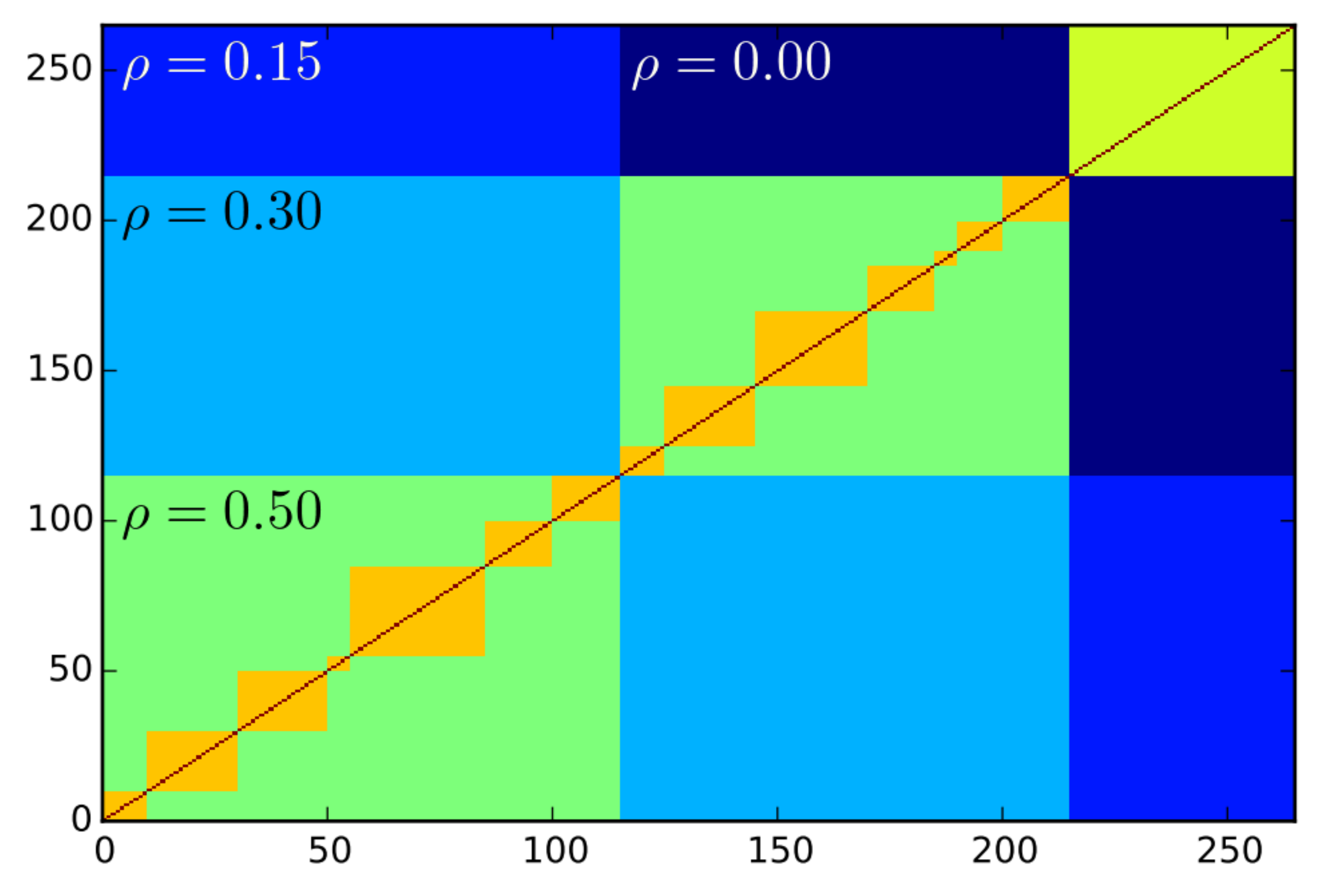}
\caption{Illustration of the correlation structure used for simulations: European assets (numbered $0,\ldots,214$) are subdivided into $2$ clusters which are themselves subdivided into $7$ clusters each; Japanese assets (numbered $215\ldots,264$) are weakly correlated to the European markets: $\rho = 0.15$ with ``investment grade'' assets, $\rho = 0.00$ with ``high-yield'' assets}\label{fig:correl_mat}
\end{figure}

The Gaussian and $t$-distribution are parameterized by a covariance matrix $\Sigma$.
We define $\Sigma$ such that the underlying correlation matrix has the structure depicted in Figure~\ref{fig:correl_mat}. This correlation structure is inspired by the real correlations between credit default swap assets in the European ``investment grade'', European ``high-yield'' and Japanese markets. 
More precisely, this correlation matrix allows us to simulate the returns time series for $N = 265$ assets divided into
\begin{itemize}
\item a ``European investment grade'' cluster composed of $115$ assets, subdivided into
\begin{itemize}
\item $7$ industry-specific clusters of sizes 10, 20, 20, 5, 30, 15, 15; the pairwise correlation inside these $7$ clusters is $0.7$;
\end{itemize}
\item a ``European high-yield'' cluster composed of $100$ assets, subdivided into
\begin{itemize}
\item $7$ industry-specific clusters of sizes 10, 20, 25, 15, 5, 10, 15; the pairwise correlation inside these $7$ clusters is $0.7$;
\end{itemize}
\item a ``Japanese'' cluster composed of $50$ assets whose pairwise correlation is $0.6$.
\end{itemize}

We can then sample time series from these two models.

\subsection{Experiment: Recovering the initial clusters}

\begin{figure*}
\begin{center}
\includegraphics[width=0.33\linewidth]{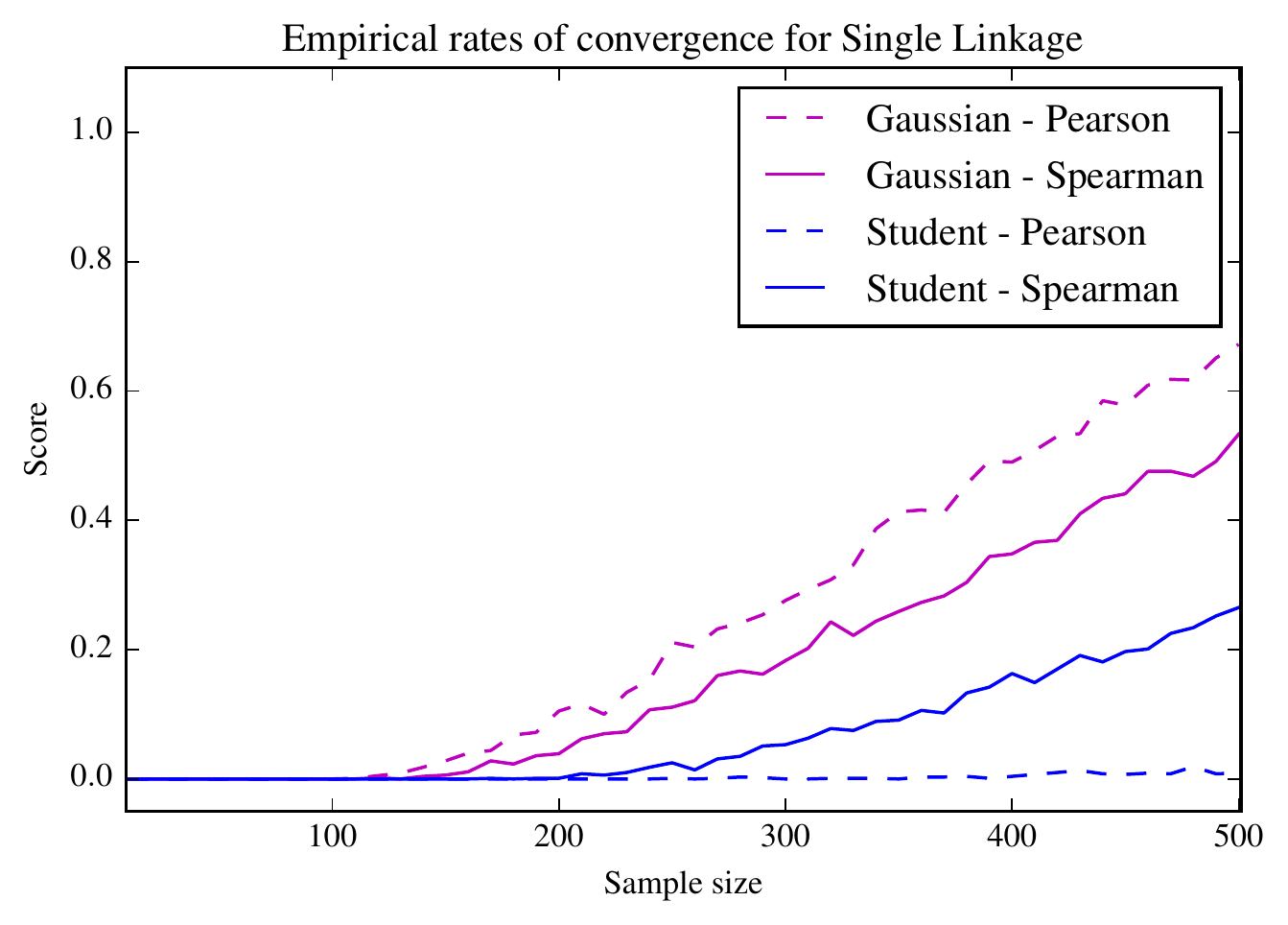}
\includegraphics[width=0.33\linewidth]{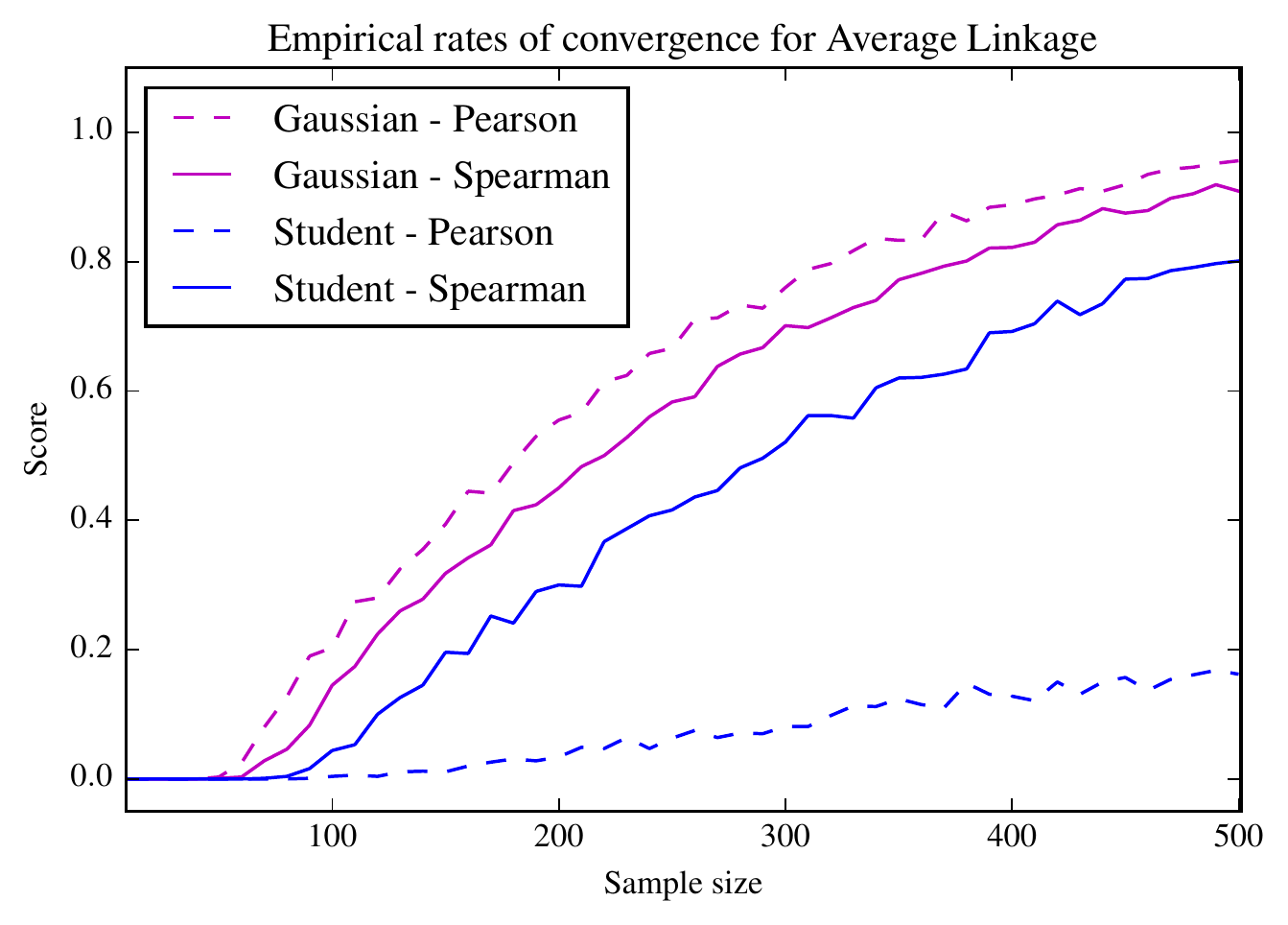}
\includegraphics[width=0.33\linewidth]{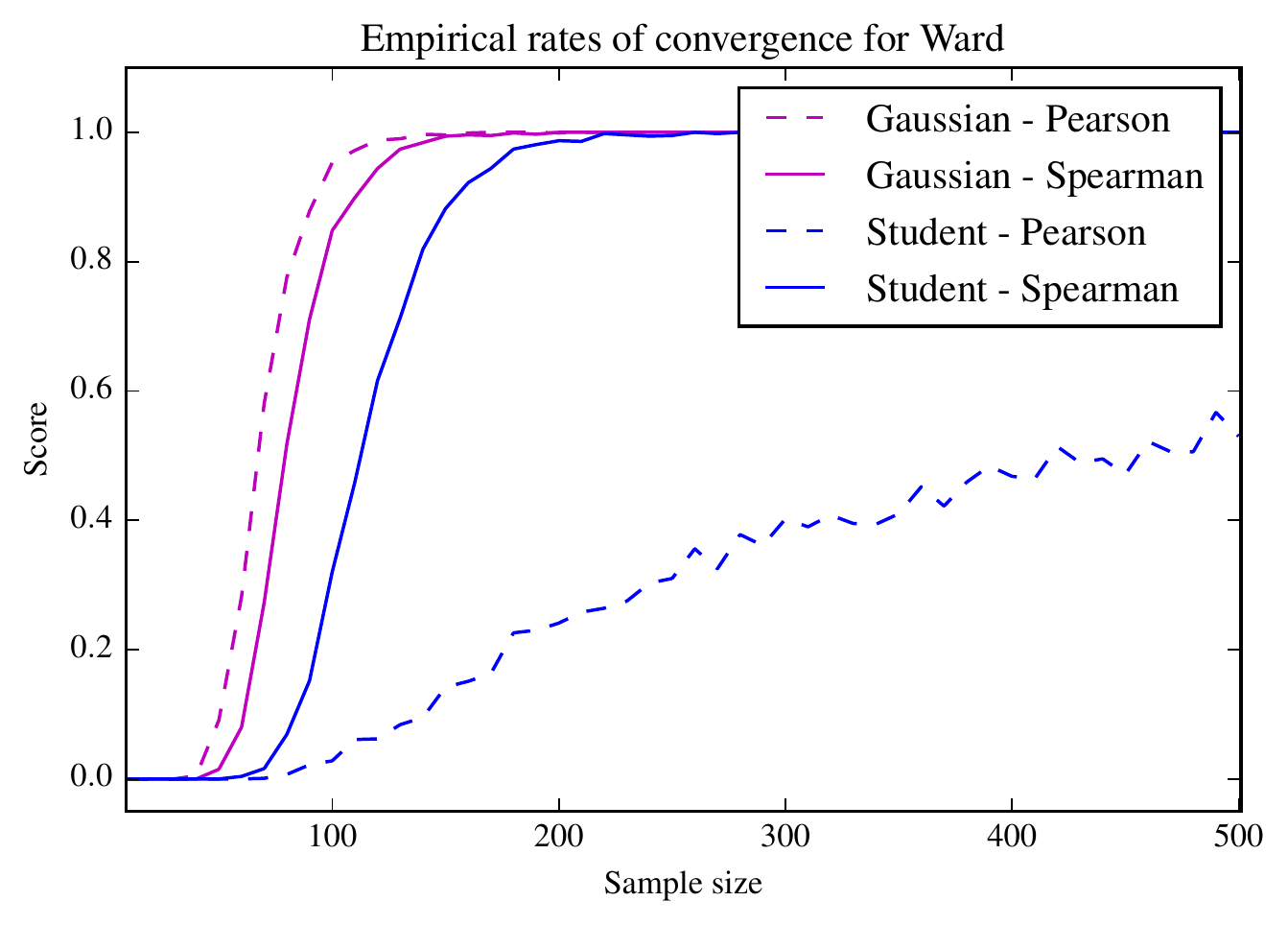}
\end{center}
\caption{Single Linkage (left), Average Linkage (mid), Ward method (right) are used for clustering the simulated time series; Dashed lines represent the ratio of correct clustering over the number of trials when using Pearson coefficient, solid lines for the Spearman one; Magenta lines are used when the underlying model is Gaussian, blue lines for the $t$-distribution}\label{fig:conv_rates}
\end{figure*}

For each model, for every $T$ ranging from $10$ to $500$, we sample $L = 10^3$ datasets of $N = 265$ time series with length $T$ from the model. We count how many times the clustering methodology (here, the choice of an algorithm and a correlation coefficient) is able to recover the underlying clusters defined by the correlation matrix. In Figure~\ref{fig:conv_rates}, we display the results obtained using \textit{Single Linkage} (motivated in Mantegna \textit{et al.}'s research \cite{mantegna1999introduction} by the ultrametric space hypothesis and the related subdominant ultrametric given by the minimum spanning tree), \textit{Average Linkage} (which is used to palliate against the unbalanced effect of Single Linkage, yet unlike Single Linkage, it is sensitive to monotone transformations of the distances $d_{ij}$) and the \textit{Ward method} leveraging either the Pearson correlation coefficient or the Spearman one.

\subsection{Conclusions from the empirical study}

As expected, the Pearson coefficient yields the best results when the underlying distribution is Gaussian and the worst when the underlying distribution is heavy-tailed. For such elliptical distributions, rank-based correlation estimators are more relevant \cite{liu2012transelliptical,han2013optimal}.
Concerning clustering algorithm convergence rates, 
we find that Average Linkage outperforms Single Linkage for $T \ll N$ and $T \simeq N$. One can also notice that both Single Linkage and Average Linkage have not yet converged after 500 realizations (roughly 2 years of daily returns) whereas the Ward method, which is not mainstream in the econophysics literature, has converged after 250 realizations (about a year of daily returns). Its variance is also much smaller. Based on this empirical study, a practitioner working with $N = 265$ assets whose underlying correlation matrix may be similar to the one depicted in Figure~\ref{fig:correl_mat} should use the Ward + Spearman methodology on a sliding window of length $T = 250$.

\section{Discussion}

In this contribution, we only show consistency with respect to a model motivated by empirical evidence.
\textit{All models are wrong} and this one is no exception to the rule: random walk hypothesis, real correlation matrices are not that ``blocky". We identified several theoretical directions for the future:
\begin{itemize}
\item The theoretical concentration bounds are not sharp enough for usual values of $N, T$. Since the intrinsic dimension of the correlation matrices in the HCBM is low, there might be some possible improvements \cite{tropp2015introduction}.
\item ``Space-conserving", ``space-dilating'' is a coarse classification that does not allow to distinguish between several algorithms with different behaviors. Though Single Linkage (which is nearly ``space-contracting'') and Average Linkage have different convergence rates as shown by the empirical study, they share the same theoretical bounds.
\end{itemize}

\begin{figure}
\includegraphics[width=\columnwidth]{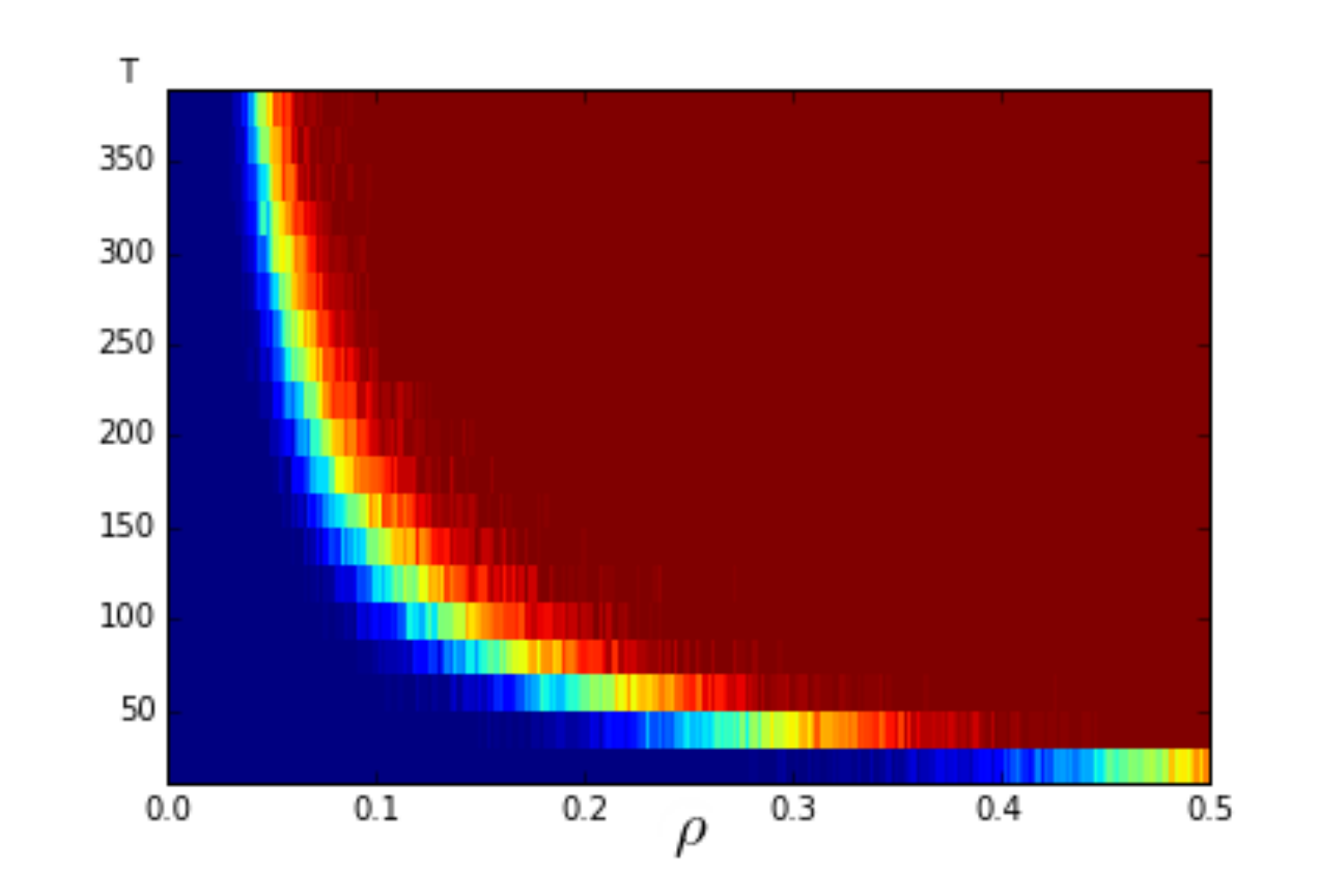}
\caption{Heatmap encoding the ratio of correct clustering over the number of trials for the Ward + Spearman methodology as a function of $\rho$ and $T$; underlying model is a Gaussian distribution parameterized by a 2-block-uniform-$\rho$ correlation matrix; red color represents a perfect and systematic recovering of the underlying two clusters, deep blue encodes 0 correct clustering; notice the clear-cut isoquants}
\label{fig:isoquant}
\end{figure}

And also directions for experimental studies:
\begin{itemize}
\item It would be interesting to study spectral clustering techniques which are less greedy than the hierarchical clustering algorithms. In \cite{tumminello2007kullback}, authors show that they are less stable with respect to statistical uncertainty than hierarchical clustering. Less stability may imply a slower convergence rate.
\item We notice that there are isoquants of clustering accuracy for many sets of parameters, e.g. $(N,T)$, $(\rho,T)$. Such isoquants are displayed in Figure~\ref{fig:isoquant}. Further work may aim at characterizing these curves. We can also observe in Figure~\ref{fig:isoquant} that for $\rho \leq 0.08$, the critical value for $T$ explodes. It would be interesting to determine this asymptotics as $\rho$ tends to 0.
\end{itemize}

Finally, we have provided a guideline to help the practitioner set the critical window-size $T$ for a given clustering methodology. One can also investigate which consistent methodology provides the correct clustering the fastest. However, much work remains to understand the convergence behaviors of clustering algorithms on financial time series.

\bibliographystyle{named}
\bibliography{ijcai16}

\end{document}